\documentclass{article}
\usepackage[T1]{fontenc} %

\usepackage[preprint]{corl_2026} %

\usepackage{graphicx}
\usepackage{booktabs}
\usepackage{amsmath}
\usepackage{amssymb}
\usepackage{xcolor}
\usepackage{pgfplots}
\pgfplotsset{compat=1.18}
\usetikzlibrary{calc,arrows.meta}
\usepackage{tabularx}
\usepackage{enumitem}
\usepackage{caption}
\usepackage{float}
\newif\ifshowtodos\showtodosfalse

\newcommand{\projecturl}{\url{https://zero-real.github.io/}}

\newcommand{\zap}{ZAP}
\newcommand{\astra}{GPT-6 Astra}

\title{Pretraining Knows What, Simulation Teaches How: Compositional Generalization for Agentic Robotics}

\author{
  \textbf{Benjamin Pikus, Daniel Saunders, Mircea Mironenco,}\\
  \textbf{Salman Mohammadi, Mathew Mathew, Sirou Zhu, Divya Makkar,}\\
  \textbf{Xuhua Huang, Qinbo Li, Supreeth Narasimhaswamy,}\\
  \textbf{Siyuan Cen, Dejia Xu$^*$, Shyamal Buch$^*$, Thomas Neff\thanks{equal advising}}\\
  Luma AI \\
  \projecturl
}

\hypersetup{pdftitle={Pretraining Knows What, Simulation Teaches How: Compositional Generalization for Agentic Robotics},
  pdfauthor={Benjamin Pikus, Daniel Saunders, Mircea Mironenco, Salman Mohammadi, Mathew Mathew, Sirou Zhu, Divya Makkar, Xuhua Huang, Qinbo Li, Supreeth Narasimhaswamy, Siyuan Cen, Dejia Xu, Shyamal Buch, Thomas Neff}}

\begin{document}
\maketitle

\begin{abstract}
Frontier models such as \astra{} now directly control robots and can generalize better than vision-language-action models (VLAs).
We argue that this ability could arise from two distinct, composable skills: \textbf{semantic reasoning}, to understand what a task asks for in a scene, and \textbf{motion control}, to move the robot to do it. We show that small open VLMs are already capable
of the first and mainly lack the second.
We introduce \textbf{Zero-real Agent Policy (\zap{})}, a recipe that teaches this missing motion skill, entirely in simulation on simple primitive tasks and from the model's own successes, with \textit{no frontier teacher or distillation} and \textit{no real-world robot data}.
\zap{} reaches \textbf{96.7\%} success on real-world tasks, matching \astra{} in our harness while running 3$\times$ faster and 17$\times$ cheaper, and \textbf{74.0\%} on a bimanual adaptation of RoboLab-120, on par with \astra{}'s 72.2\%. \zap{} also generalizes to new objects, goals, and strategies, and in simulation to a robot embodiment on which it never trained.
\end{abstract}

\begin{figure}[h]
  \centering
  \input{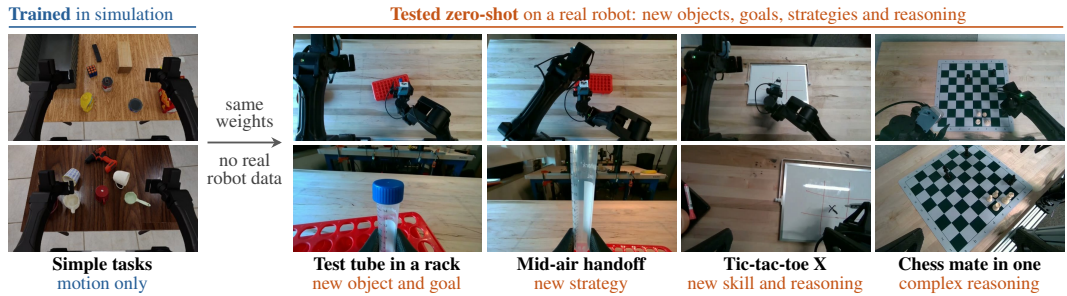}
  \caption{\textbf{ZAP teaches small open VLMs the missing motion skill without frontier model distillation.} \textbf{Left:} \zap{} trains on simple simulated tasks. \textbf{Right:} \zap{} generalizes to unseen \textit{real-world} tasks with novel objects, goals, and complex strategies not present in training (e.g., test tube, mid-air handoff between arms, tic-tac-toe, and chess mate-in-one), without any real robot data.}
  \label{fig:teaser}
\end{figure}

\section{Introduction}
\label{sec:intro}

\astra{}~\citep{openai2026astra} has changed what the robotics community expects from frontier vision-language models. Controlling a robot through a tool harness, it significantly outperforms state-of-the-art vision-language-action (VLAs) and world-action models
(WAMs)~\citep{yang2026astranomy,zhang2026unexpected}, and many now expect that robot policies will need frontier-scale models~\citep{williams2026astrashock}. Yet a trillion-parameter model is slow and costs dollars per task: e.g., in our harness, \astra{} takes 397\,s and \$4.87 on an average real-world task. We ask the question: \textit{is it possible that a much smaller VLM already has most of what is needed, provided it is given the right factorization and post-training?}

We first factorize robot agent policies into two fundamental, compositional skills. 
\textbf{Semantic reasoning} is the capability to understand what a task asks for in a scene, e.g., which objects are \emph{behind}, precisely counting objects, or finding the mate-in-one in chess.
\textbf{Motion control} instructs a robot how to move: how to grasp, how to approach the target at the right height, angle and velocity, and translating 3D perception inputs into poses the robot can reach.

Given this factorization, we then investigate \textit{to what extent a small VLM can already perform each of these two skills}. With our harness~(\S\ref{sec:harness}), we find that while \astra{} succeeds on 96.7\% of real-world trials, base Qwen 3.6 27B succeeds on just 16.7\%. 
Looking deeper at the planning calls and reasoning, we observe that the small VLM consistently understands the high level goal and scene, but fails to select the correct grasp, does not approach the target correctly, and cannot translate its 3D perception into motion commands~(\S\ref{sec:diagnosis}). %
Thus, it is clear that VLM pretraining brings sufficient \textit{semantic reasoning} (i.e., understanding the ``what'' of the task and scene), but is unable to perform the critical \textit{motion control} reasoning (i.e., understanding the ``how'' of robotic execution) required to output an effective agentic robot policy. 

We introduce \textbf{Zero-real Agent Policy (\zap{})}, which adds the missing motion control capability to a pretrained vision-language model, allowing it to move the robot on a wide range of tasks.
\zap{} uses a minimal harness to control end-effector poses and gripper openings, without grasp macros or motion primitives, and uses the same harness across simulation and real-world hardware.
We first fine-tune the model on its own successful rollouts and then continue training via GRPO, following a dense reward ladder to facilitate full task completions.
With as few as 40 simulated tasks that are \emph{deliberately simple}, and \emph{without a frontier teacher model} or \emph{real-world robot data}, we demonstrate that \zap{} can combine reasoning from pretraining with motion from simulation on tasks that appear in neither.
Compared to prior work that distills from frontier-model trajectories~\citep{liu2026guava,li2026racap,li2026roboharness}, \zap{} is not bounded by the capabilities of a teacher model.

Trained only on simple simulated pick and place, stacking and reorientation tasks, we show that \zap{} generalizes to \textit{unseen tasks in real-world settings} and \textit{unseen embodiments}. 
In a real-world setting, \zap{} seats a test tube vertically in a rack, hands it between arms in mid-air when asked, draws an X on a tic-tac-toe board, and plays a mate-in-one in chess (Figure~\ref{fig:teaser}). 
\zap{} matches \astra{} at 3$\times$ the speed and 1/17 the cost. 
On RoboLab-120~\citep{yang2026robolab} it is on par with \astra{} and ahead of Opus 5.5, and on a simulated Franka arm it never trained on, it outperforms the leading VLA and WAM policies. Our contributions are:
\begin{itemize}[leftmargin=*,itemsep=1pt,topsep=2pt]
  \item \textbf{Teaching motion control.} 
  We show that a small open VLM (Qwen 3.6 27B) already understands robot tasks and mainly lacks motion control, with only 36\% of its motion calls correct in our harness (\S\ref{sec:diagnosis}). 
  \zap{} teaches only motion control, in simulation, on simple tasks and from the model's own successes, with no frontier teacher and no real robot data (\S\ref{sec:method}).
  \item \textbf{Frontier-level performance at a fraction of the cost.} 
  We show that \zap{} can match \astra{} on real hardware at 3$\times$ the speed and 1/17 the cost, and on RoboLab-120 achieves comparable scores while clearly outperforming top VLAs (\S\ref{sec:quant}).
  \item \textbf{Compositional generalization.} 
  \zap{} handles new objects, goals and strategies on real tasks never seen in training, such as a mid-air handoff, tic-tac-toe and chess (\S\ref{sec:compose}). 
  Without retraining, the same checkpoint also transfers to a single-arm Franka in simulation, reaching 55.9\% on RoboLab-120, ahead of the leading VLA and WAM policies (\S\ref{sec:embodiment}).
\end{itemize}

\section{Related work}
\label{sec:related_main}

\textbf{Foundation models as robot agents.} 
Code as Policies and its successors have a language model write programs over perception and control APIs~\citep{liang2022code,huang2023voxposer,fu2026capx}. 
Frontier models now also act as closed-loop policies through tool harnesses~\citep{jia2026agentpolicy,robocurve2026inspect,menon2026astraarms}, and \astra{} in particular outperforms VLAs and WAMs~\citep{yang2026astranomy,zhang2026unexpected}. 
Robo-Harness K1 suggests that a capable VLM needs additional 3D perception and exposes it as tools~\citep{li2026roboharness}, and Show-Harness proposes a semantic action interface~\citep{chen2026showharness}. 
Even when providing 3D object measurements via our harness to the model, we still find that small models lack motion control, so \zap{} targets this weakness in training.

\textbf{Compact trained robot agents.} 
Guava, RACaP and Robo-Harness K1 train small tool-calling agents by fine-tuning on trajectories from a frontier teacher~\citep{liu2026guava,li2026racap,li2026roboharness}.
Show-Harness fine-tunes on human and agent demonstrations~\citep{chen2026showharness}, and CaP-RL trains a code-writing model with RL in simulation~\citep{fu2026capx}. 
Since \zap{} learns from its own rollouts, it is not limited by the capabilities of a teacher. 
It controls end-effector poses instead of using grasp tools or semantic action units, and generalizes to new real-world tasks and, in simulation, to a robot it never trained on.

\textbf{End-to-end policies.} 
VLAs and WAMs learn reasoning and motion together from robot action data~\citep{brohan2023rt2,black2024pi0,ye2026dreamzero,fang2026molmoact2}. 
They keep the intent of an instruction but fail to execute it on new tasks or skill combinations~\citep{zhou2025agnostos,fang2025intact,zhou2025liberopro}, and they struggle with instructions about how to carry out a task~\citep{park2026instructions}. 
\zap{} only learns motion from robot interaction, and its language understanding comes from pretraining, so it does not depend on the instructions contained within the robot data.

\textbf{Separating reasoning from motion.} 
Task-and-motion planners compose symbolic operators that ground into geometric motion~\citep{toussaint2015lgp,garrett2021tamp}, and language models have since served as high-level planners over libraries of learned skills~\citep{ahn2022saycan,huang2022innermonologue,lin2023text2motion} or low-level policies~\citep{dalal2024psl,dalal2025manipgen}. 
\zap{} keeps the split, but builds motion control into the model itself rather than offloading it to a static library.

Appendix~\ref{sec:related} gives a more detailed discussion.

\section{Method}
\label{sec:method}

\begin{figure}[t]
  \centering
  \input{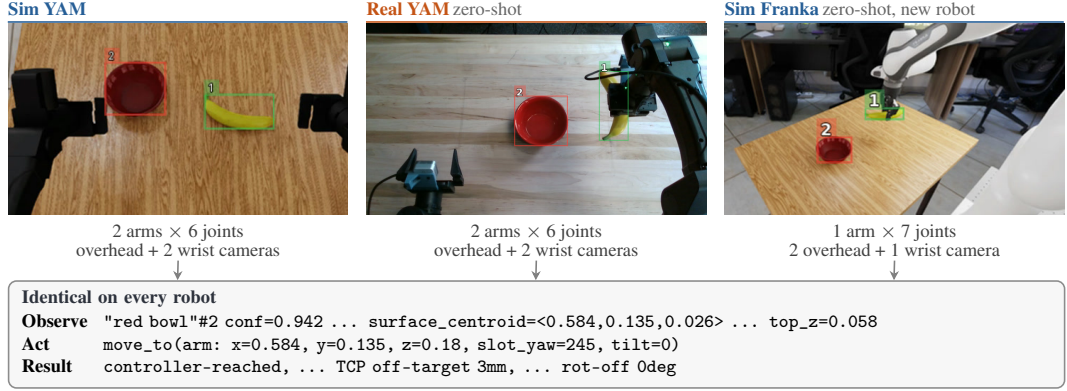}
  \caption{\textbf{Three scenes, one shared interface.} Left to right: separate banana-in-bowl episodes on the simulated bimanual YAM, the real YAM (all real world settings are unseen during training), and a simulated single-arm Franka (an unseen embodiment). From the agent's perspective, the only thing that changes is the system prompt specifying the details of the robot and camera layouts. Across all three settings, \zap{} uses the \textit{same} interface: camera images annotated with SAM~3.1 masks and numbered boxes, 3D object readouts, end-effector pose commands, and controller feedback. The box below shows one step from the Franka episode: a 3D object readout, the agent's \texttt{move\_to} command, and the motion controller's response. Both YAM settings use these same formats.}
  \label{fig:method}
\end{figure}

We first describe our harness through which the model perceives and acts, which is identical in simulation and on hardware (\S\ref{sec:harness}).
We then investigate the base model's capabilities in our harness (\S\ref{sec:diagnosis}), and teach it the missing motion control in simulation, first by fine-tuning on its own successes and then with GRPO and a dense reward ladder (\S\ref{sec:training}).

\subsection{The \zap{} harness}
\label{sec:harness}

The \zap{} harness connects the model to a robot through tool calls and is the same in simulation and on hardware (Figure~\ref{fig:method}). 
Our main robot is a bimanual rig with two 6-DoF (degrees-of-freedom) YAM arms and parallel-jaw grippers~\citep{i2rt_yam}. 
Through the harness, the model controls absolute end-effector poses and gripper openings. 
We frame tasks as a conversation between the model and its environment through the harness: the model first writes a plan and names the objects to locate based on camera views of the environment, then repeatedly issues perception and movement tool calls to complete its task.

Our harness uses SAM~3.1~\citep{meta2026sam31,carion2025sam3} to segment the objects named in the model's perception queries in each camera stream. 
These masks are combined with calibrated depth cameras to yield each object's 3D centroid, height, and principal axes.
This process runs continuously and asynchronously in the background: the model steers perception by changing its queries, and receives object measurements and annotated camera images. 

The \zap{} harness provides five tools: set end-effector poses for one or both arms (\texttt{move\_to}), set gripper openings (\texttt{set\_gripper}), refine perception targets (\texttt{change\_query}), keep a short plan in view across turns (\texttt{update\_todo}), and end the episode (\texttt{done}). 
We do not include grasp macros or motion primitives, since we found that they raise the floor but also cap the ceiling (Appendix~\ref{app:harness}). 

Appendix~\ref{app:harness} describes further harness design decisions: 3D pose and orientation parameterization, the observations, validation, the motion controller and its outcome reports, and how the same harness runs in simulation and on hardware. 
Appendix~\ref{app:episode} shows a full real-robot episode turn by turn.

\subsection{Pretraining knows what, but not how}
\label{sec:diagnosis}
In our harness, base Qwen 3.6 27B succeeds on 8.3\% of RoboLab-120 scenes and 16.7\% of real-world trials.
An episode can either fail by misunderstanding the task or by wrong motion control, so we investigate both by looking at the RoboLab-120 episodes.

\textbf{Task understanding.} 
In 95\% of scenes, the base model's plan correctly identifies target objects, destination and object order. 
In 70\% of scenes, the perception query correctly returns every target and destination object. 
But even in episodes where both cases hold, the base model still only successfully completes the task 9\% of the time.

\textbf{Motion control.} 
We find that only 36\% of the base model's motion calls are correct (Figure~\ref{fig:motion-bars}). 
32\% are refused by the controller because the requested pose is physically impossible, 15\% are misdirected (the right intent but the wrong pose, such as a grasp that stops above the object), and 16\% are pointless, such as carrying an empty gripper.
Thus, the base model's failures mostly stem from incorrect motion control. 
We close that gap by teaching motion control in simulation (\S\ref{sec:training}), and later verify that semantic reasoning does not degrade with improved motion control (\S\ref{sec:motion-not-meaning}).

\begin{figure}[t]
    \centering
    \input{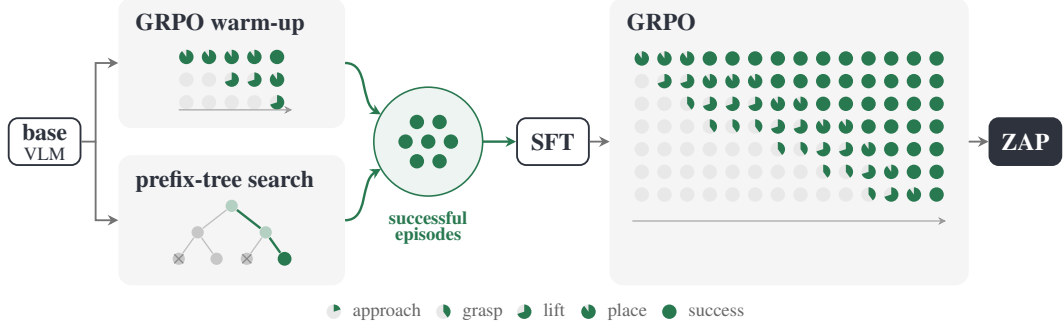}
    \caption{\textbf{How \zap{} learns motion.} Starting from the base VLM, a short GRPO warm-up and a prefix-tree search over partial episodes each yield successful episodes on the simulated tasks. Those episodes seed an SFT stage, and a much longer GRPO run with a dense per-episode reward (partial credit for approaching, grasping, lifting and placing; full credit for success) produces \zap{}. \zap{} is trained exclusively in simulation.}
    \label{fig:pipeline}
\end{figure}
  
\subsection{Simulation teaches how}
\label{sec:training}
\label{sec:init}

We build our simulated tasks in Isaac Sim~\citep{nvidia_isaacsim} by adding simulated YAM arms with calibrated cameras and arm dynamics against real YAM arms.

\textbf{Learning motion control on simulated tasks.} 
We design 40 simulated tasks, none of which appear in RoboLab-120, that exercise everyday object manipulation skills: grasping and lifting, single- and multi-object pick and place, stacking, and reorientation. 
The goals are deliberately simple, and do not require deep reasoning, which lets us test whether motion control learned on simple tasks can combine with pretrained semantic reasoning to solve unseen tasks with new goals and strategies.

\textbf{Seeding the model with its own successes.} 
A model which is on-distribution for its task space learns more effectively during RL training, so we first fine-tune on successful agent traces. 
One approach is to rejection sample from a fixed base model or larger open model. However, this is prohibitively inefficient, and many tasks never succeed even with hundreds of attempts. We opt to run a short GRPO warm-up to mine successful traces.
We run GRPO with several dozen steps, over-sampling the more difficult tasks, and collect successful rollouts for a subsequent SFT stage.
As the policy improves, harder tasks start to yield successes, so the data comes from a slightly evolved version of the model itself, a self-distillation loop that is close to on-policy (Figure~\ref{fig:pipeline}).

\textbf{Prefix-tree search.} 
For tasks that never succeed, even after the GRPO warm-up, we perform a tree-search over partial episodes (Appendix Figure~\ref{fig:tree}). 
We add a node each time a \texttt{move\_to} or \texttt{set\_gripper} call has finished and the scene has come to rest.
To expand the node, we reset the simulator to the task's deterministic start, replay the node's stored actions, restore the agent's context, and sample a new continuation. 
To pick which node to expand, we score each one by how often its branches have succeeded, task progress, and how rarely the node has been tried, and we prune branches that keep failing.
This found successful traces for several tasks that RL sampling never solved, and these traces are added to the fine-tuning data.

\textbf{GRPO with dense rewards.} The GRPO run is initialized from this SFT checkpoint and trains on all 40 tasks (hyperparameters in Appendix~\ref{app:hparams}). Binary reward based on task success is inefficient since some tasks start with zero successes, and therefore a pass/fail reward would provide zero advantage and contribute no gradient. We instead shape a dense ladder of partial credit judged on the final state of the scene, manually designed per task type. 

For example, for pick and place tasks, approaching the object earns up to 0.2 in proportion to the gripper fingertip-to-object distance closed; grasping earns 0.2 and lifting up to a further 0.55, in proportion to height up to 8\,cm. Transport raises the total to at most 0.9, in proportion to the distance to the goal closed. Completing a task earns 5.0, minus a small length penalty which penalizes the model for taking too many turns to complete the task. The specific numbers for partial credit were not ablated and instead holistically chosen: our guiding principle was that final task success must have a much higher reward than partial reward; otherwise, the model settles for collecting partial credit. The success checker is hardened against reward hacking, for example, by requiring that the goal hold for at least half a second.

We penalize turns where either no tool call is made, or where invalid motion is proposed and doesn't pass certain validation checks (see Appendix~\ref{app:harness}). These turns (but not the whole sequence) receive a negative advantage, regardless of the episode's score. In earlier runs without these penalties, the corresponding error rates reached as high as 4\% and 34\%, respectively. With penalties, these rates fall to 0\% and 3\%. 

\section{Experiments}
\label{sec:quant}

We compare \zap{} with frontier models in simulation and on real hardware (\S\ref{sec:frontier}), check what training actually taught the VLM (\S\ref{sec:motion-not-meaning}), test it on new real tasks (\S\ref{sec:compose}) and on a new embodiment (\S\ref{sec:embodiment}), and ablate the recipe (\S\ref{sec:support}).

\subsection{\zap{} matches frontier models at a fraction of the cost}
\label{sec:frontier}
\label{sec:sim}
\label{sec:real}

\begin{table}[!htbp]
  \centering
  \small
  \begin{tabular}{@{}lc@{}}
    \toprule
    Model & Success (\%) \\
    \midrule
    Base Qwen 3.6 27B & 8.3 \\
    \zap{} (SFT) & 58.3 \\
    \zap{} (SFT + RL) & \textbf{74.0} \\
    \midrule
    \astra{} & 72.2 \\
    Opus 5.5 & 61.2 \\
    \bottomrule
  \end{tabular}
  \caption{\textbf{Success on the bimanual YAM adaptation of RoboLab-120}~\citep{yang2026robolab}. \zap{} is on par with \astra{} and ahead of Opus 5.5~\citep{anthropic2026opus55}.}
  \label{tab:sim}
\end{table}

\textbf{RoboLab-120 in simulation.} On the bimanual YAM adaptation of RoboLab-120, \zap{} is on par with \astra{} and ahead of Opus 5.5 (Table~\ref{tab:sim}).

\textbf{Real-world.} We recreate 10 tasks of diverse complexity from the RoboLab-120 task list on the real robot (Appendix~\ref{app:pertask} lists each task prompt and each model's success rate). \zap{} matches \astra{} (reasoning level max) in the same harness, each succeeding on 29 of 30 trials (Table~\ref{tab:real}), and both clearly outperform \astra{} in the Robocurve harness, MolmoAct2, and the base model.

  \begin{table}[t]
    \centering
    \small
    \begin{minipage}[t]{0.47\linewidth}
      \centering
      \setlength{\tabcolsep}{3pt}
      \begin{tabular}{@{}lcc@{}}
        \toprule
        Model & Success (\%) & 95\% CI \\
        \midrule
        \textbf{\zap{} (ours, 27B)} & \textbf{96.7} & 83--99 \\
        \astra{} (\zap{} harness) & \textbf{96.7} & 83--99 \\
        \astra{} (Robocurve) & 36.7 & 22--54 \\
        MolmoAct2 & 33.3 & 19--51 \\
        Base Qwen 3.6 27B & 16.7 & 7--34 \\
        \bottomrule
      \end{tabular}
    \end{minipage}\hfill
    \begin{minipage}[t]{0.51\linewidth}
      \centering
      \begin{tabular}{@{}lcc@{}}
        \toprule
        Per task & \zap{} & \astra{} \\
        \midrule
        Time (s) & \textbf{146} $\pm$ 83 & 397 $\pm$ 159 \\
        Turns & 21.9 $\pm$ 10.9 & 19.4 $\pm$ 8.0 \\
        Time per turn (s) & \textbf{6.2} $\pm$ 4.5 & 19.0 $\pm$ 8.6 \\
        Cost (USD) & \textbf{0.29} $\pm$ 0.16 & 4.87 $\pm$ 3.03 \\
        \bottomrule
      \end{tabular}
    \end{minipage}
    \caption{\textbf{Real-world results.} Left: success rate (\%) over 30 trials per model, with 95\% Wilson intervals. Right: mean $\pm$ std per task. \zap{} is served with vLLM~\citep{kwon2023vllm} on one B300 priced at \$7.10/hour, with a 2-token multi-token-prediction head.}
    \label{tab:real}
  \end{table}

\textbf{VLA limitations.} 
MolmoAct2 shows the opposite pattern to the base VLM: it can execute the required motions on these tasks, but struggles with semantic reasoning (Appendix Table~\ref{tab:pertask}). For example, we evaluate the model's ability to pick \& place a Rubik's cube in different locations. It places a Rubik's cube in a bowl in all three trials, but fails all three when asked to place it \emph{behind} the bowl, often putting it inside instead. When asked to place exactly three bananas in a bin and no more, it consistently adds too many. Although VLAs also initialize from VLMs, their ability to turn concepts such as \emph{behind} or \emph{exactly three} into actions is constrained by the coverage of their robot training data.

\textbf{3$\times$ faster, 17$\times$ cheaper.} \zap{} and \astra{} solve these tasks the same way. They take a similar number of turns, choose the same target objects, destinations and ordering, and make the same kinds of motion calls. Since \zap{} is significantly smaller, \zap{} finishes a task in 146 seconds for \$0.29, while \astra{} takes 397 seconds and \$4.87 (Table~\ref{tab:real}).

\subsection{Training changes motion, not understanding}
\label{sec:motion-not-meaning}

\S\ref{sec:diagnosis} identified motion control as the base model's main weakness. We now compare the base model with \zap{} to examine what training changed.

\textbf{Semantic reasoning remains the same.}
If training had changed semantic reasoning, base and \zap{} would disagree on goals and perception. They do not.
The two models agree on the target objects, destination, and object order in more than 90\% of scenes. Similarly, their perception calls return the same objects in 70\% of scenes. Even when we restrict the comparison to episodes with matching perception, the gap in task success remains large: 9.0\% for the base model versus 73.6\% for \zap{}. %

\begin{figure}[t]
\centering
\includegraphics[width=\linewidth]{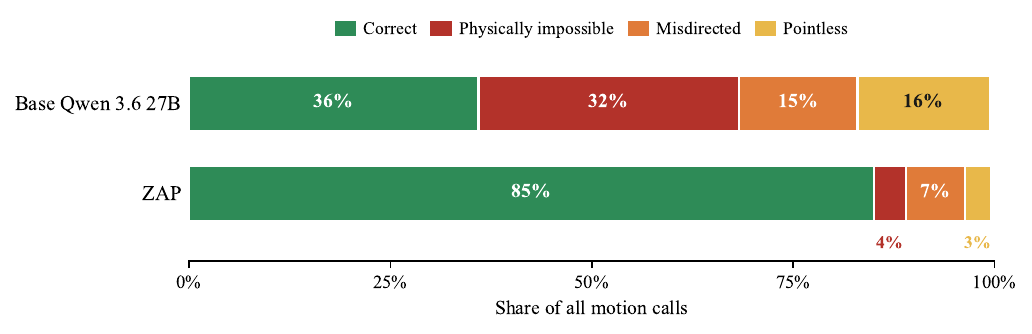}
\caption{\textbf{Taxonomy of Base and \zap{}'s motion.} Every move command issued by both models sorted into four categories: correct, physically impossible, executed with the right intent but the wrong pose, such as grasping too high above the object (misdirected), or executed on a false premise, such as carrying an empty gripper (pointless).}
\label{fig:motion-bars}
\end{figure}

\begin{figure}[t]
\centering
\includegraphics[width=\linewidth]{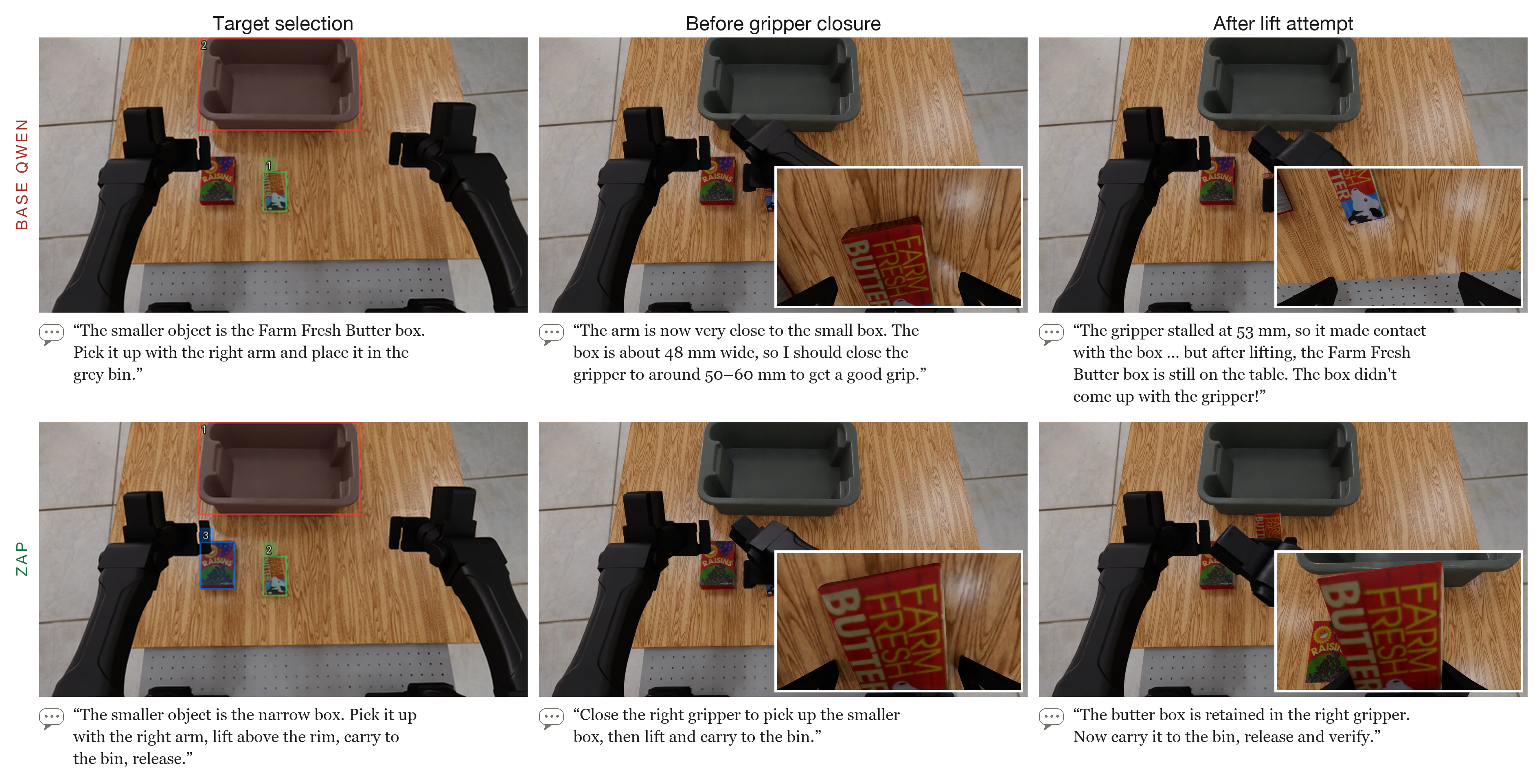}
\caption{\textbf{Different motion example.} The task is to place the smaller object in the gray bin on the same scene. Below each image is the summarized version of the model's own reasoning. Both identify the butter box and choose the right arm. However, the base model stops 0.6\,cm above the box, closes the jaw, and lifts nothing. \zap{} lowers the fingertips to the table on either side of the box, closes until the jaws stop against it, and lifts it toward the bin.}
\label{fig:grasp-episode}
\end{figure}

\textbf{The difference is motion control.} 
85\% of \zap{}'s motion calls are correct, against 36\% of the base model's (Figure~\ref{fig:motion-bars}). Figure~\ref{fig:grasp-episode} shows one such episode: both models identify the butter box and plan the same pick-and-place, but the base model cannot figure out the right grasp approach.

\subsection{Composing the skills on new real tasks}
\label{sec:compose}

We next test \zap{}, trained exclusively in simulation, on real-world tasks far outside anything seen in training, from bimanual handoffs to drawing and chess, each requiring reasoning and motion the model never practiced (Figure~\ref{fig:teaser}).

\textbf{Generalization axes.} We consider the following axes for generalization in our analysis:
\begin{itemize}[leftmargin=*,itemsep=1pt,topsep=2pt]
  \item \emph{New tasks and objects:} the policy should handle objects, scenes, and goals that appear nowhere in training, and construct a physical strategy for them without demonstrations.
  \item \emph{New strategies on request:} it should be possible to prompt the policy into manipulations it never practiced, such as coordinating two arms.
  \item \emph{Complex reasoning:} the policy should be able to carry over the reasoning capabilities the pretrained VLM has into physical world actions.
  \item \emph{Language and instruction following:} unlike VLAs, the policy should be highly steerable by its prompt, and not sensitive to slight deviations in wording when the meaning is the same.
\end{itemize}

Importantly, \textit{none} of the goals, objects, or scenes below appear in training, and neither do any of the more complex prompts, strategies, or reasoning they require.

\begin{figure}[!htbp]
  \centering
  \setlength{\tabcolsep}{1pt}
  \begin{tabular}{@{}cccc@{}}
    \includegraphics[width=0.245\linewidth]{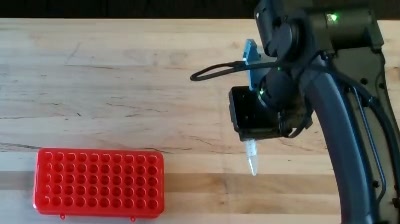} &
    \includegraphics[width=0.245\linewidth]{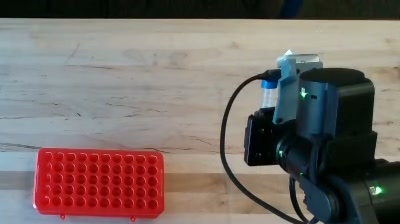} &
    \includegraphics[width=0.245\linewidth]{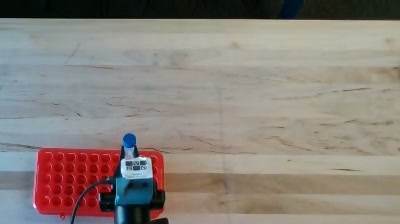} &
    \includegraphics[width=0.245\linewidth]{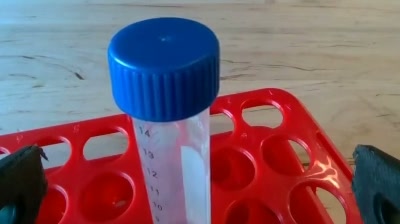} \\
    \scriptsize (a) Pick up the tube & \scriptsize (b) Rotate the gripper & \scriptsize (c) Carry it over the rack & \scriptsize (d) Set it into a hole
  \end{tabular}
  \caption{\textbf{A new object and a new goal.} Frames from the single test tube episode (a--c: overhead camera, zoomed; d: wrist camera). The model picks up the tube lying on its side, rotates the gripper so the tube stands upright, carries it over the rack and lowers it into a hole, cap up.}
  \label{fig:tube}
\end{figure}

\begin{figure}[!htbp]
  \centering
  \setlength{\tabcolsep}{1pt}
  \begin{tabular}{@{}cccc@{}}
    \includegraphics[width=0.245\linewidth]{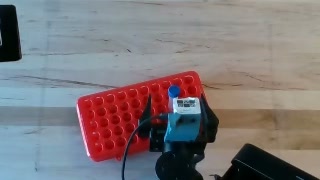} &
    \includegraphics[width=0.245\linewidth]{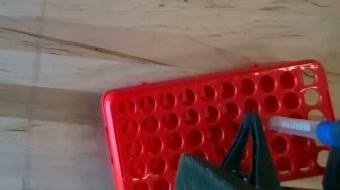} &
    \includegraphics[width=0.245\linewidth]{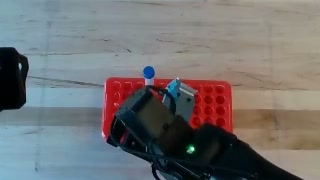} &
    \includegraphics[width=0.245\linewidth]{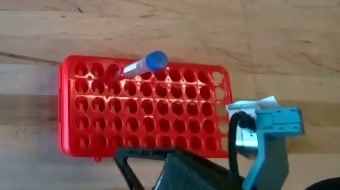} \\
    \multicolumn{2}{c}{\scriptsize (a) Default prompt: overhead and wrist views} & \multicolumn{2}{c}{\scriptsize (b) ``Top left of the rack'': overhead and wrist views}
  \end{tabular}
  \caption{\textbf{Steered by the instruction.} Final scenes of two episodes from the same starting scene: (a) the default prompt; (b) a request for a hole at the top left of the rack.}
  \label{fig:steer}
\end{figure}

\textbf{A new object and a new task: the test tube.} We ask the model to vertically lower a test tube into a rack (Figure~\ref{fig:tube}). Both objects and the goal are unseen in training. The model constructs the strategy itself: it grasps the tube lying on its side, turns the gripper so the tube is vertical, and lowers it into a hole. This is challenging because, at the moment of placement, the hole is hidden from every camera. A VLA would typically need task-specific demonstrations of each of these steps; \zap{} receives only a single text instruction and, without demonstrations or parameter updates, infers the required motion.

\textbf{Steered by the instruction.} The model brings the language understanding of its pretrained LLM into its actions. VLAs often respond poorly to changes in their prompts~\citep{watts2026rephrase}. \zap{} is able to follow precise instructions; on the same starting scene, we replace the instruction with a request to place the tube in a hole at the top left of the rack. With the default prompt, the model picks a hole on its own; with the new one, the tube ends up in the top-left corner (Figure~\ref{fig:steer}).

\begin{figure}[!htbp]
  \centering
  \setlength{\tabcolsep}{1pt}
  \begin{tabular}{@{}ccccc@{}}
    \includegraphics[width=0.196\linewidth]{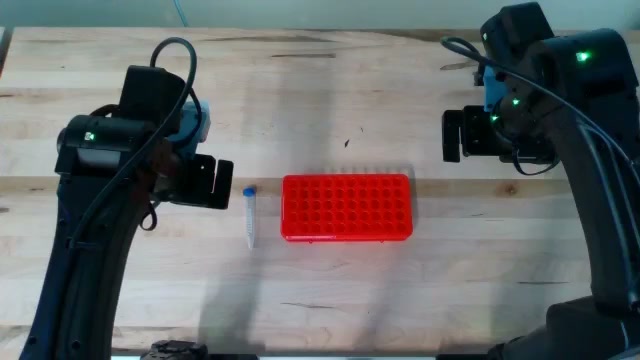} &
    \includegraphics[width=0.196\linewidth]{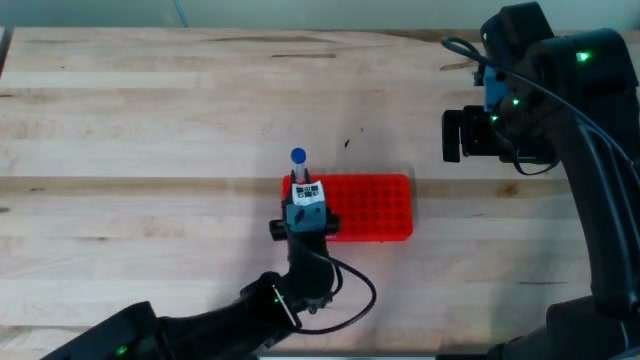} &
    \includegraphics[width=0.196\linewidth]{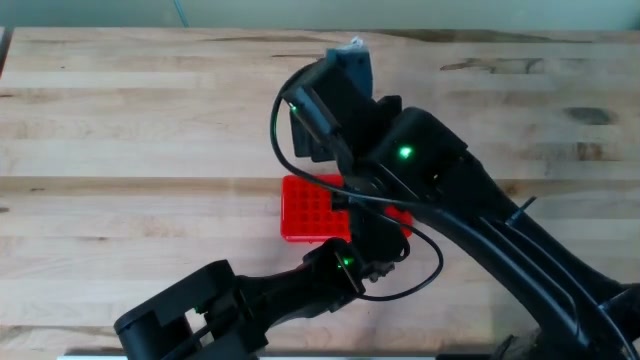} &
    \includegraphics[width=0.196\linewidth]{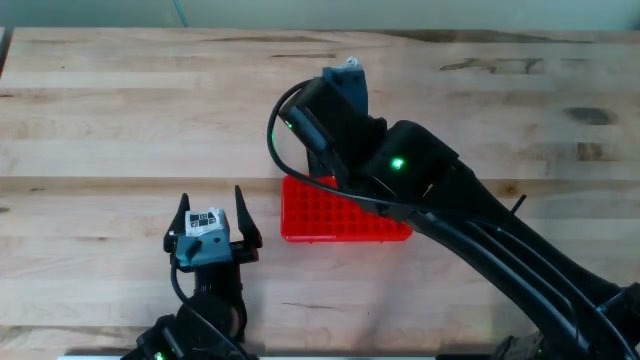} &
    \includegraphics[width=0.196\linewidth]{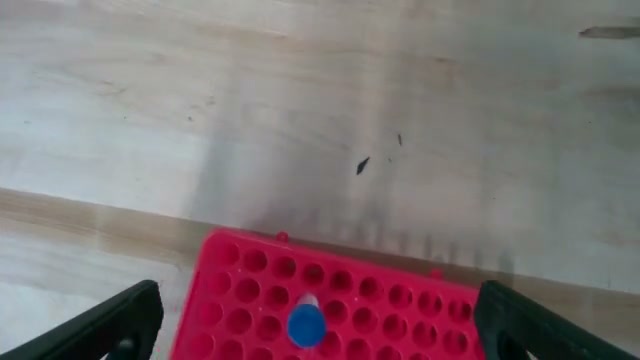} \\
    \scriptsize (a) Start & \scriptsize (b) Held up mid-table & \scriptsize (c) Other arm grasps cap & \scriptsize (d) Giving arm away & \scriptsize (e) Placed (wrist)
  \end{tabular}
  \caption{\textbf{Mid-air bimanual handoff.} Frames from the handoff episode, described step by step in the text. Training contains no bimanual interaction close to a handoff.}
  \label{fig:handoff}
\end{figure}

\textbf{A new strategy on request: bimanual handoff.} We then ask for a completely different way to reach the same goal: hand the tube from one arm to the other in mid-air, then place it in the rack (Figure~\ref{fig:handoff}). The prompt describes the handoff in writing, but training contains no bimanual interaction even close to a handoff. The model orients the tube cap-up and brings it, held by one arm, to the middle of the table; brings the other arm over the tube and descends from above, just far enough to grasp the cap but not so low that it hits the first arm; closes the receiving gripper, then opens the giving one; moves the giving arm out of the way; and lowers the tube into a hole. Getting this right means timing the two grippers against each other: if the giving arm lets go too early the tube falls, and if the receiving arm comes in too low the grippers collide.

\begin{figure}[!htbp]
  \centering
  \setlength{\tabcolsep}{1pt}
  \begin{tabular}{@{}cccc@{}}
    \includegraphics[width=0.245\linewidth]{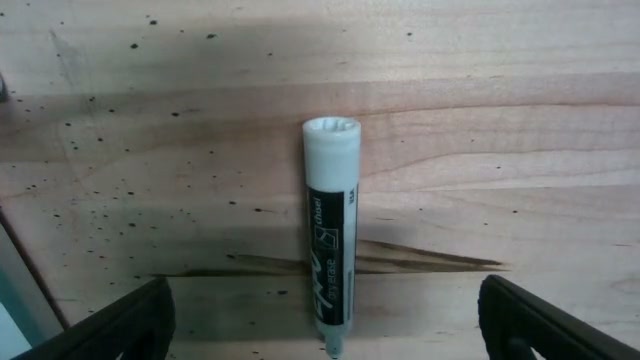} &
    \includegraphics[width=0.245\linewidth]{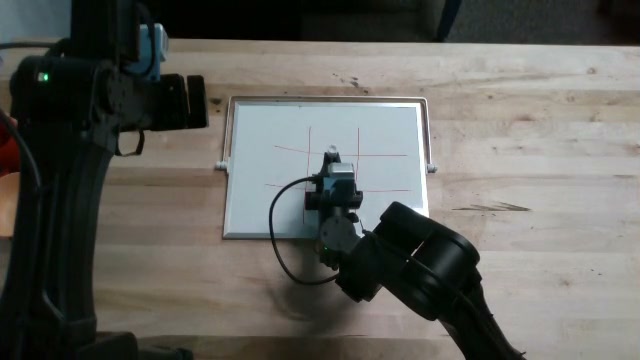} &
    \includegraphics[width=0.245\linewidth]{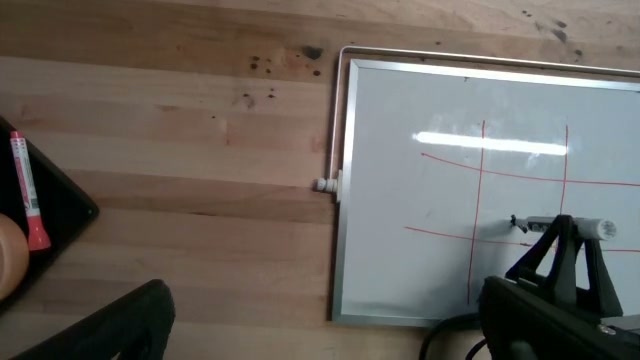} &
    \includegraphics[width=0.245\linewidth]{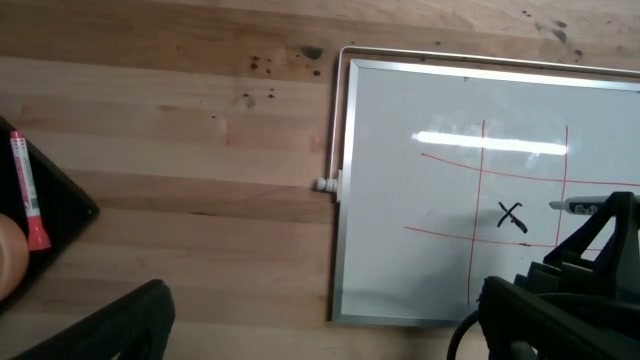} \\
    \scriptsize (a) Picking up the marker & \scriptsize (b) Marker over the center & \scriptsize (c) First stroke & \scriptsize (d) X in the center square \\[3pt]
    \includegraphics[width=0.245\linewidth]{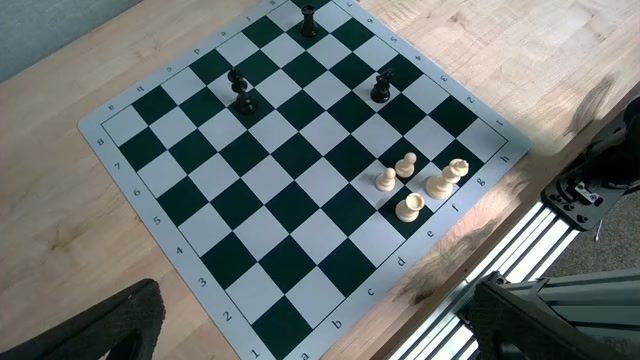} &
    \includegraphics[width=0.245\linewidth]{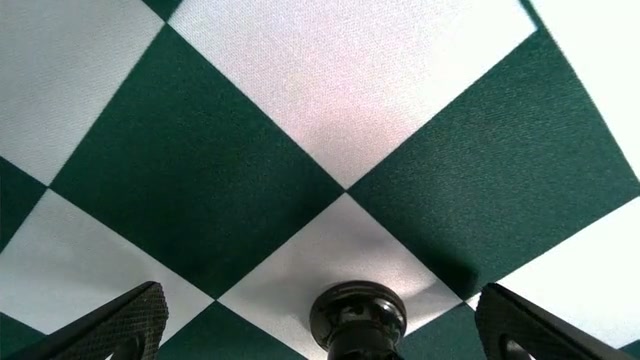} &
    \includegraphics[width=0.245\linewidth]{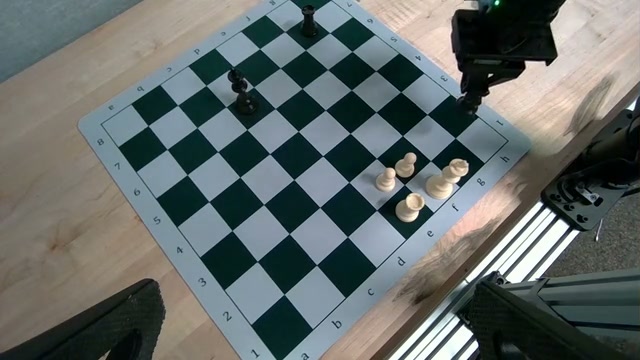} &
    \includegraphics[width=0.245\linewidth]{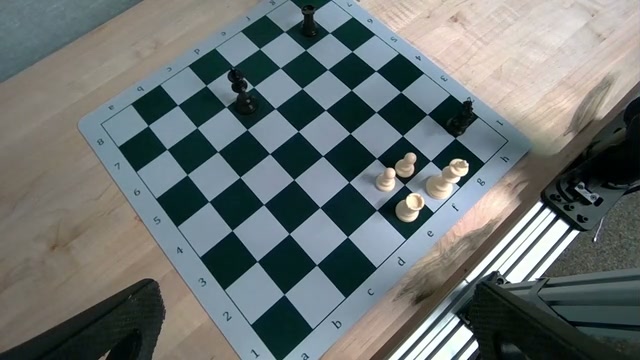} \\
    \scriptsize (e) Starting position & \scriptsize (f) Piece grasped (wrist view) & \scriptsize (g) Carrying the piece & \scriptsize (h) Checkmate played
  \end{tabular}
  \caption{\textbf{Reasoning past what the scene shows.} Top: tic-tac-toe, prompted to draw an X in the center of the board. Bottom: chess, prompted with black to move.}
  \label{fig:games}
\end{figure}

\textbf{Complex reasoning: tic-tac-toe and chess.} The last two tasks require the model to reason about something broader than object placement (Figure~\ref{fig:games}). First, we prompt the model to draw an X in the center of a tic-tac-toe game drawn on a whiteboard. It picks up the marker, recognizes what a tic-tac-toe board is and where its center square lies, and draws the X: it keeps enough contact to leave a line through each stroke, without pressing so hard that it breaks or dislodges the marker.

Second, we prompt the model to make a chess move, with black to move. It observes the board, reasons about the rules of chess, finds the mating move for black, and then picks up the correct piece and sets it on the target square to complete the checkmate. This requires the composition of reasoning from the VLM and object manipulation from simulation training; neither skill alone would have produced the checkmate.

\subsection{Transfer to a new embodiment}
\label{sec:embodiment}

\textbf{Same checkpoint, new robot.} We take the same checkpoint, trained only on bimanual YAM arms, and evaluate it in simulation on a single-arm Franka. From the agent's perspective, only the system prompt describing the embodiment changes; the harness stays exactly the same. The Franka differs from the YAM in most of the ways that matter to a policy that outputs joint angles or action chunks: one arm instead of two, seven joints instead of six, a different gripper, different reach, and different cameras. \zap{} is unaffected by these differences: it commands end-effector poses through the harness, which handles the embodiment-specific joint control.

\begin{table}[!htbp]
  \centering
  \small
  \begin{tabular}{@{}lc@{}}
    \toprule
    RoboLab-120, single-arm Franka & Success (\%) \\
    \midrule
    \textbf{\zap{} (transferred from bimanual YAM)} &  \textbf{55.9} \\
    Lakeside & 51.0 \\
    FLUX 3 Action & 42.9 \\
    \bottomrule
  \end{tabular}
  \caption{\textbf{The same motion skill on a new robot.} Without any Franka training, \zap{} is ahead of Lakeside and FLUX 3 Action, the leading VLA/WAM policies on the current RoboLab-120 leaderboard~\citep{nvidia2026robolableaderboard}.}
  \label{tab:embodiment}
\end{table}

On RoboLab-120, \zap{} on this unseen embodiment still reaches 55.9\%, ahead of Lakeside (51\%) and FLUX 3 Action~\citep{bfl2026flux3action} (42.9\%), the leading VLA/WAM policies on the RoboLab-120 leaderboard~\citep{nvidia2026robolableaderboard} (Table~\ref{tab:embodiment}). What the model learned in simulation is therefore not a mapping tied to the YAM arms, but how to manipulate the robot arm through space to move an object to a target. The harness lets that skill carry over from two YAM arms to one Franka without retraining.

\subsection{Ablations}
\label{sec:support}

\textbf{Better SFT data improves generalization.} We compare two SFT runs: one using 455 successful traces mined early in GRPO, and one using 3.9k traces mined more aggressively after GRPO runs longer, especially on harder tasks. RoboLab-120 success rises from 17.2\% to 58.3\% (Table~\ref{tab:support}, left), a gain of 41.1 points from the same recipe with better data. %

\textbf{The recipe also works when trained directly on Franka.} Separately, we train Qwen 3.5 9B on single-arm Isaac Sim tasks. As before, none of these tasks appear in the RoboLab-120 evaluation. We use a smaller 9B model both for faster experiments, and to test whether the recipe also holds at a smaller scale. We follow a similar, lighter version of the recipe as above; we first train on 4.6k examples gathered from mining successes from GRPO runs, before performing a final GRPO run. The policy exceeds current SOTA VLAs (Table~\ref{tab:support}, right), suggesting that the recipe works across model sizes and embodiments while preserving generalization to entirely unseen tasks. 

\begin{table}[!htbp]
  \centering
  \small
  \begin{minipage}[t]{0.49\linewidth}
    \centering
    \begin{tabular}{@{}lcc@{}}
      \toprule
      SFT data & Traces & Success (\%) \\
      \midrule
      Early mining & 455 & 17.2 \\
      Longer mining & 3.9k & \textbf{58.3} \\
      \bottomrule
    \end{tabular}
  \end{minipage}\hfill
  \begin{minipage}[t]{0.47\linewidth}
    \centering
    \begin{tabular}{@{}lc@{}}
      \toprule
      Franka-trained model & Success (\%) \\
      \midrule
      \zap{} 9B & \textbf{53.4} \\
      \midrule
      Lakeside & 51.0 \\
      FLUX 3 Action & 42.9 \\
      \bottomrule
    \end{tabular}
  \end{minipage}
  \caption{\textbf{Ablations.} Left: more extensive trace mining improves SFT generalization on RoboLab-120 with the bimanual YAM. Right: a 9B model trained directly on separate single-arm Isaac Sim tasks also transfers to RoboLab-120. Lakeside and FLUX 3 Action results are from the benchmark leaderboard~\citep{nvidia2026robolableaderboard}.}
  \label{tab:support}
\end{table}

\section{Conclusion}
\label{sec:conclusion}

In this work, we demonstrated how small-scale VLMs can output agentic robotics policies comparable to frontier models by considering the right factorization and post-training. We factorized robotics policies into two compositional skills, \textit{semantic reasoning} and \textit{motion control}. In our harness, we found that a representative small VLM (base Qwen 3.6 27B) plans the right goal in 95\% of RoboLab-120 scenes but succeeds in only 8.3\% of them, because most of its motion calls are refused, misdirected or pointless. We then introduced \zap{}, which trains that missing motion control reasoning skill, on 40 simple simulated tasks and bootstrapped from the model's own successes, with \textit{no frontier teacher or distillation and no real robot data}. The resulting 27B policy matches \astra{} at 3$\times$ the speed and 1/17 the cost, reaches 74.0\% on bimanual RoboLab-120 against \astra{}'s 72.2\%, and transfers in simulation to an unseen single-arm Franka embodiment. On real tasks well outside its training, from seating a test tube to a mid-air handoff and a chess mate-in-one, it combines reasoning from pretraining with motion from simulation; we show these as qualitative examples. Adding a skill to \zap{} means writing new simulated tasks, not collecting new robot demonstrations.

\clearpage

\bibliography{refs}

\clearpage
\appendix

\section{The harness}
\label{app:harness}

\textbf{Tools.} \zap{} has five tools. \texttt{move\_to} moves one or both arms to a complete end-effector pose; naming both arms in one call produces a single synchronized motion. \texttt{set\_gripper} sets each jaw opening, from 0 (closed) to 1 (fully open). \texttt{change\_query} sets the text descriptions of the objects that perception segments. \texttt{update\_todo} writes a short plan that is repeated in every later observation, so it survives when old turns are
dropped from context. \texttt{done} ends the episode.

\textbf{Poses.} Each \texttt{move\_to} gives an absolute position of the tool center point, in meters, and a full orientation, in a world frame shared by both arms. Perception reports object positions in the same frame and units, so a measured centroid can be used directly as a target. The harness has no relative moves, grasp macros or approach primitives. In earlier versions, such macros raised success on the objects they were designed around but got in the way on everything else. Without them, the model plans the approach, grasp, lift, transport and release itself, and the motions it learned on pick and place carry over to rotating a tube, a mid-air handoff and drawing with a marker (\S\ref{sec:compose}).

\begin{figure}[H]
  \centering
  \includegraphics[width=\linewidth]{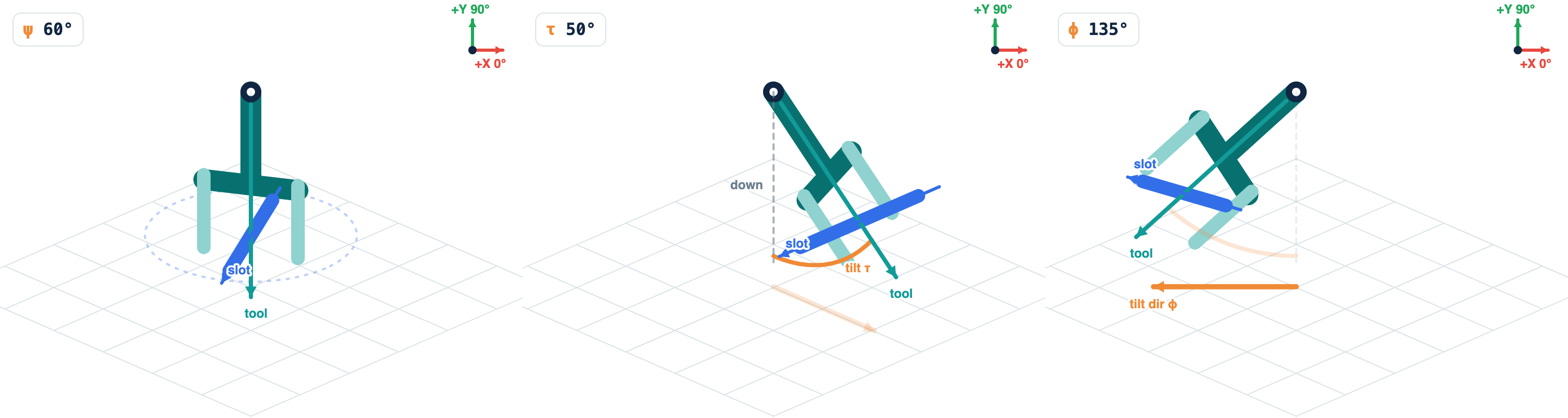}
  \caption{\textbf{Slot yaw, tilt and tilt direction.} The gripper orientation parameterization used by \texttt{move\_to} (\S\ref{sec:harness}). Left: slot yaw $\psi$ turns the jaw slot (blue), the axis a held object lies along, about the vertical. Middle: tilt $\tau$ leans the tool axis (teal) away from straight down; $0^\circ$ is top-down and $90^\circ$ horizontal. Right: tilt direction $\phi$ sets the compass direction the fingertips tip toward. Each angle changes exactly one thing about the gripper that can be seen. All angles are measured counterclockwise from world $+x$.}
  \label{fig:orientation}
\end{figure}

\textbf{Orientation.} Orientation is given by three angles, each measured counterclockwise from world $+x$ (Figure~\ref{fig:orientation}). \emph{Slot yaw} $\psi$ is the heading of the jaw slot, the axis a held object lies along. \emph{Tilt} $\tau$ is the angle of the tool axis from straight down: $0^\circ$ is top-down and $90^\circ$ is horizontal. \emph{Tilt direction} $\phi$ is the compass direction the fingertips point toward. We use these instead of Euler angles or quaternions for two reasons. Neither of those states directly where the gripper points or how its jaws are aligned, the two quantities that decide a grasp. And a horizontal tool, as in a side grasp, lies at an Euler singularity, where a few degrees of physical error can show up as a change of about $100^\circ$ in the reported angles. Our angles are ambiguous only when the tool points straight down or straight up, where tilt direction has no physical meaning.

\textbf{Observations and perception.} Each turn adds one observation to the conversation: one downscaled RGB image from each of the three cameras (for bimanual YAM, overhead and one per wrist), the perception results, the measured tool pose and jaw opening of each arm, and the remaining step budget. Perception uses SAM~3.1~\citep{meta2026sam31,carion2025sam3}, which runs continuously in the background on every camera stream rather than as a tool the model calls and waits on. It segments the objects named by the current queries, draws their masks and numbered boxes on the images, and uses depth to compute each object's 3D centroid, height and principal axes in world coordinates. The model controls perception only through \texttt{change\_query}, which it can call at any time, for example to find a missed object or to segment part of one, such as the cap of a tube. Detections are not fused across cameras; the model decides which ones refer to the same object. In simulation, perception runs on the rendered images and depth, as it does on the real cameras, and never reads any privileged information such as object poses that would be unavailable on hardware.

\textbf{Validation.} Before a \texttt{move\_to} call reaches the controller, the harness checks it against limits declared for the robot, such as tilt range and workspace bounds. A call that fails a check is refused rather than clipped to a nearby feasible value, and the model receives the reason as an error message so it can try something else.

\textbf{Controller and outcome reports.} The controller turns a valid pose into a straight-line Cartesian path, solves inverse kinematics densely along it, and refuses the motion if the resulting joint path cannot be validated. When the arm comes to rest, the controller reports one of five outcomes per arm: \emph{reached} (execution completed and the arm came to rest within tolerance), \emph{short} (the arm came to rest outside tolerance), \emph{incomplete} (execution began but did not finish, for example because of a stall), \emph{not attempted} (nothing moved, for example because no valid path exists) or \emph{unknown} (the outcome could not be determined). Any outcome other than \emph{reached} also reports its cause, whether it arose in planning or execution, and the measured position and orientation errors.

\textbf{One harness for simulation and hardware.} The harness is the same in simulation and on the real robot. Everything specific to a robot, such as joint limits, cameras and workspace bounds, comes from a declarative embodiment description. Moving from the simulator to the real rig, or to the Franka, means swapping that description and the low-level backend, and nothing else. This is why a policy trained only in simulation runs on the real robot without changes.

\section{A real-robot episode, turn by turn}
\label{app:episode}

To make the agent loop of \S\ref{sec:harness} concrete, we step through one complete episode of \zap{} on the real bimanual rig. The task is ``pick up both blocks''. The episode succeeds in 9 turns and 72.8\,s of wall-clock time, with 4 robot actions, 118{,}985 input tokens and 1{,}920 output tokens. Each image is the observation the agent received at the start of that turn (left wrist, overhead, right wrist); from turn 2 on it includes the SAM~3.1 masks and numbered boxes for the current perception queries, computed in the background. The overhead camera faces the arms from across the table, so its image-left is the robot's right. Below each image we give an excerpt of the model's reasoning, the single tool call it emitted, and the result the harness returned, which reaches the agent as the tool message at the start of the next turn. Reasoning is abridged with [\dots]; tool calls are verbatim.

\newcommand{\epturn}[5]{%
  \par\noindent\begin{minipage}{\linewidth}
  \textbf{Turn #1} \hfill {\scriptsize #2}\\[1pt]
  \includegraphics[width=\linewidth]{figures/episode/turn_0#1.jpg}\\[1pt]
  {\footnotesize \emph{Reasoning:} #3\par
  \emph{Tool call:} {\ttfamily\scriptsize #4}\par
  \emph{Result:} #5\par}
  \end{minipage}\par\vspace{6pt}}

\epturn{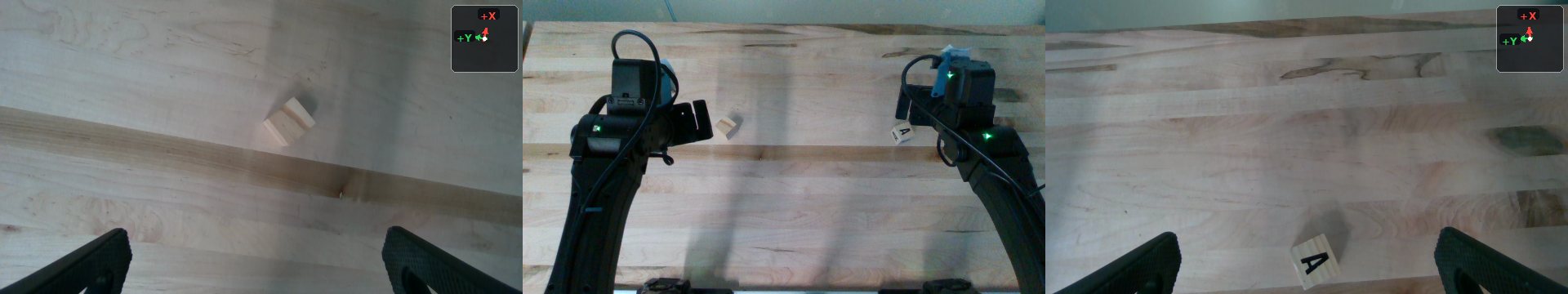}{t = 0.0--1.6\,s}{I need to pick up both blocks. I can see two blocks in the overview: one appears to be a plain wooden block on the left side, and another is a white block with the letter ``A'' on the right side. [\dots] Let me first query for both blocks to get their metric positions.}{change\_query(queries=["plain wooden block", "white block with letter A"])}{Perception query generation 1 set. The next snapshot will run these queries on its exact frame.}

\epturn{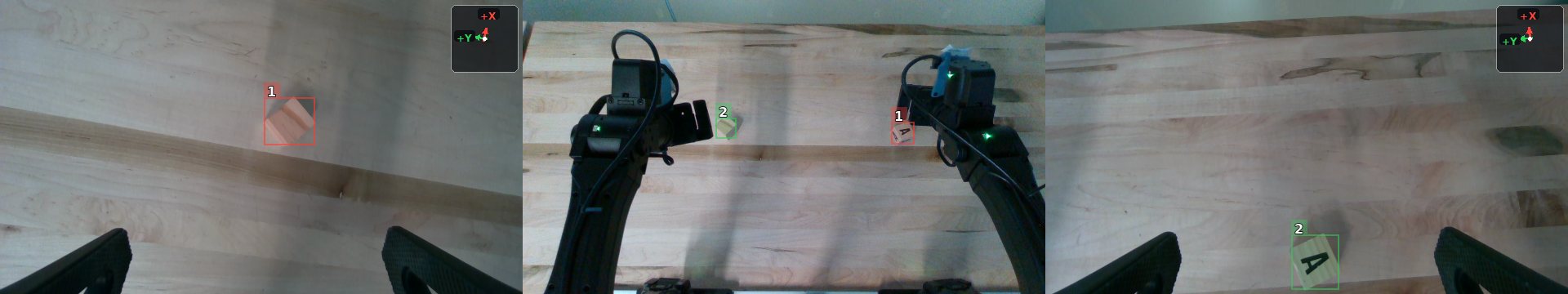}{t = 1.6--7.4\,s}{From the depth data: plain wooden block: left wrist centroid (0.412, 0.200, 0.009), top\_z 0.011, $\sim$36--39\,mm wide. A block: right wrist centroid (0.394, $-$0.237, 0.023), top\_z 0.025, $\sim$37\,mm wide. I'll use left arm for plain block and right arm for A block. [\dots] I need to approach top-down, descend, close, lift, and verify both retained.}{update\_todo(items=["Pick both blocks: plain at \textasciitilde<0.412,0.200>, top z0.011; A at \textasciitilde<0.394,-0.237>, top z0.025.", "Use left for plain, right for A. Approach top-down at z0.12, descend to grasp heights \textasciitilde0.003 and 0.010, close, lift to z0.15 and verify retention."])}{Todo list replaced (2 items).}

\epturn{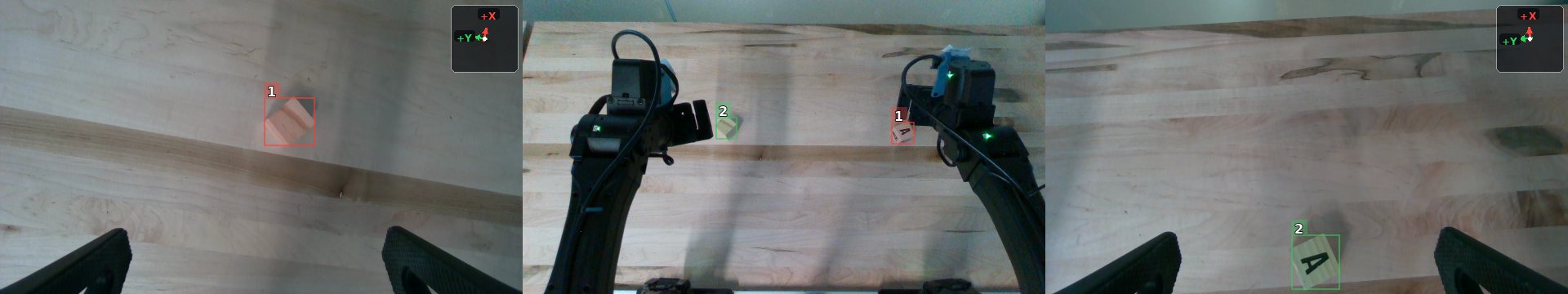}{t = 7.4--20.8\,s}{[\dots] First, I'll move both arms to approach positions above the blocks with top-down orientation.}{move\_to(left: x=0.412, y=0.2, z=0.12, slot\_yaw=180, tilt=0; right: x=0.394, y=-0.237, z=0.12, slot\_yaw=180, tilt=0)}{Both arms \emph{reached}; position error $\le$1\,mm, orientation error $<$0.2$^\circ$; inter-arm skew 15\,ms.}

\epturn{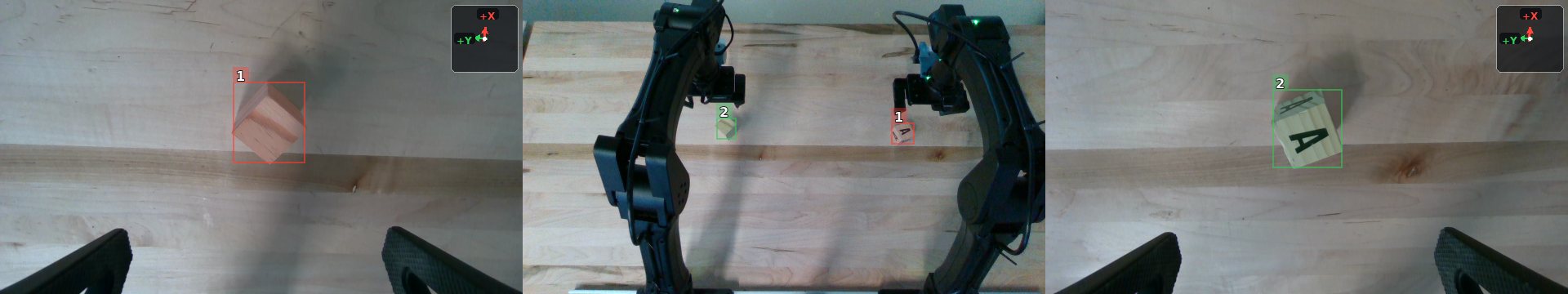}{t = 20.8--32.2\,s}{The blocks are now clearly visible in the wrist cameras. [\dots] Plain block: grasp at z0.003. A block: grasp at z0.010. I'll descend, close, lift, and verify.}{move\_to(left: x=0.412, y=0.2, z=0.003, slot\_yaw=180, tilt=0; right: x=0.394, y=-0.237, z=0.01, slot\_yaw=180, tilt=0)}{Both arms \emph{reached}.}

\epturn{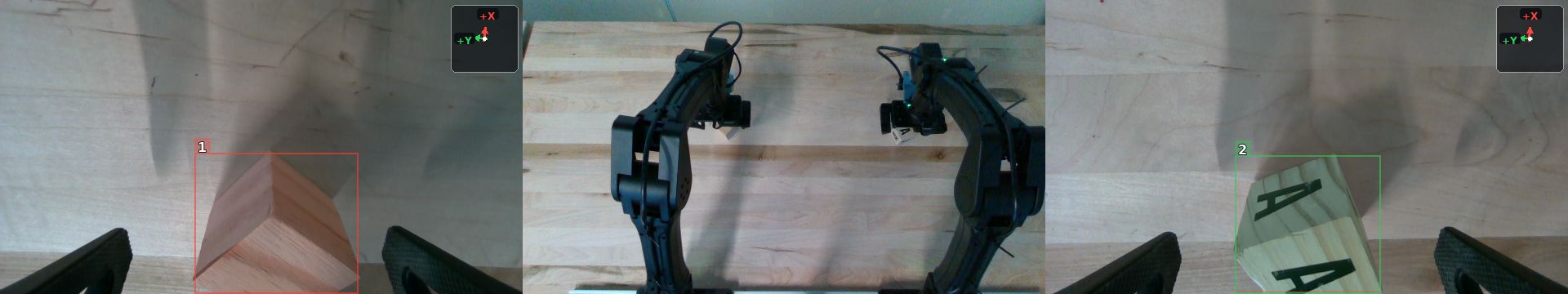}{t = 32.2--39.9\,s}{The blocks are now at the grasp heights. I need to close the grippers, lift, and verify both are retained.}{set\_gripper(left: opening=0; right: opening=0)}{Grippers \emph{stalled} as intended on the objects: left jaws stopped at 50.0\,mm, right at 37.0\,mm.}

\epturn{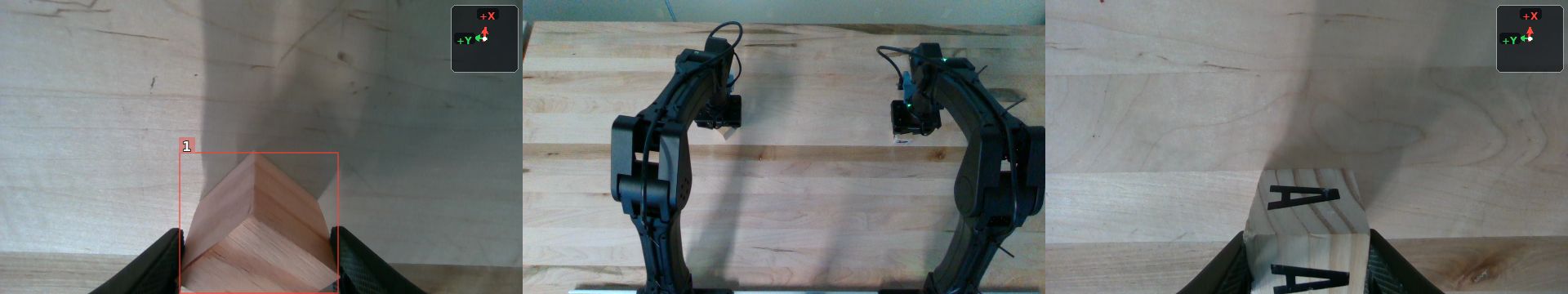}{t = 39.9--55.3\,s}{The grippers have closed on both blocks. [\dots] Lift to z0.15, keep orientation, then verify blocks are still between jaws and have moved up.}{move\_to(left: x=0.412, y=0.2, z=0.15, slot\_yaw=180, tilt=0; right: x=0.394, y=-0.237, z=0.15, slot\_yaw=180, tilt=0)}{Both arms \emph{reached}.}

\epturn{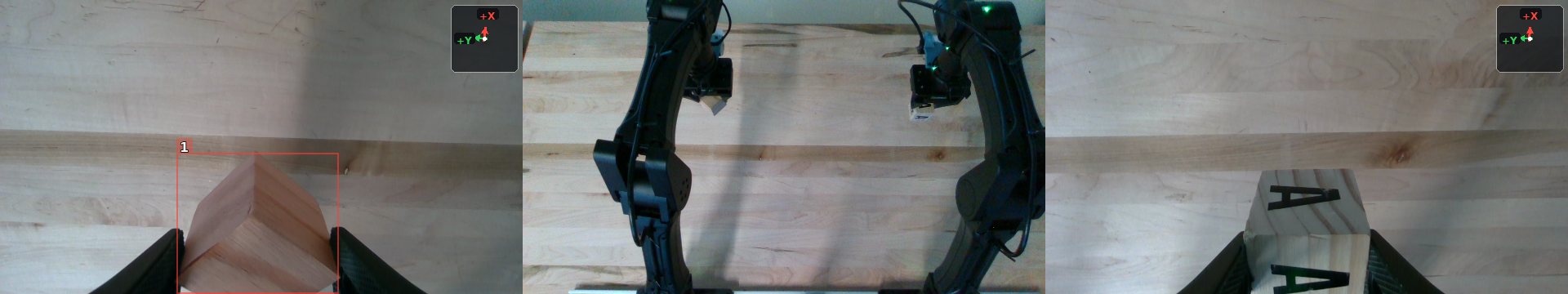}{t = 55.3--62.4\,s}{Both blocks have been lifted. [\dots] Left: plain block visible in left wrist, top\_z changed from 0.010 to 0.155 after lift, aperture 49.9\,mm. [\dots] I should verify the A block's depth after lift, and then confirm both are held.}{change\_query(queries=["wooden block with the letter A on it"])}{Perception query generation 2 set.}

\epturn{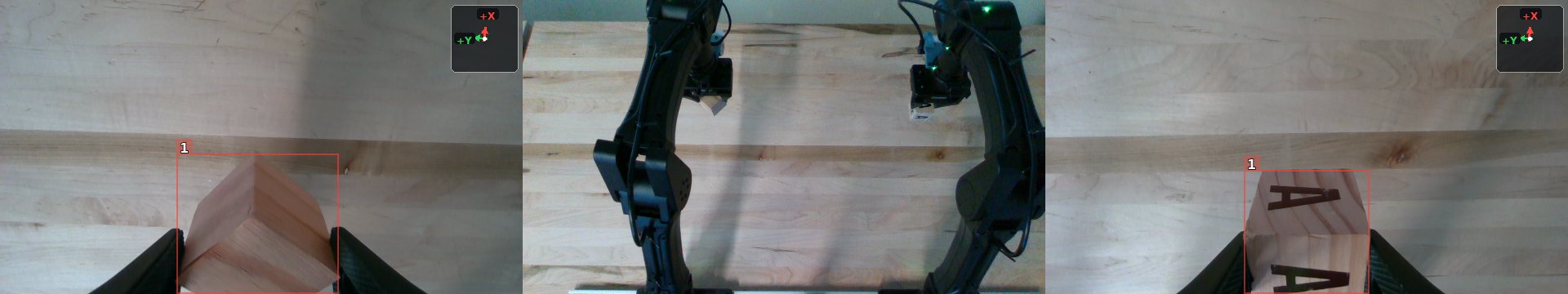}{t = 62.4--68.6\,s}{[\dots] Right: A block, initial top z0.025, now top z0.163. [\dots] Both blocks are clearly visible in the wrist cameras and their top\_z has increased from $\sim$0.01/0.025 to $\sim$0.15/0.16, confirming they are lifted off the table.}{update\_todo(items=["Both blocks successfully picked up and lifted to TCP z0.15. [\dots]", "Lift verified by block top\_z increasing [\dots] Task complete."])}{Todo list replaced (2 items).}

\epturn{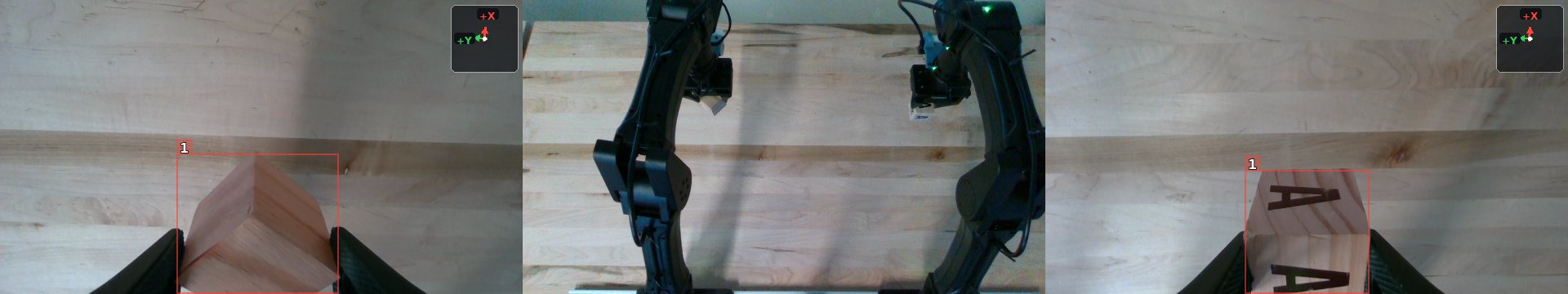}{t = 68.6--72.8\,s}{I have successfully grasped both blocks and lifted them to a height of 0.15\,m. The visual evidence confirms that both blocks are retained in the grippers after lifting. [\dots] I should now declare the task as successful.}{done(outcome="success", summary="Both blocks have been picked up. [\dots]")}{Episode ended: success.}

\section{Training details}
\label{app:hparams}

\textbf{GRPO.} We train with GRPO~\citep{shao2024deepseekmath} in the style of Dr.~GRPO~\citep{liu2025drgrpo}, with no standard-deviation or per-sequence length normalization. 2 groups per task and 8 rollouts per group: 640 episodes per batch over 40 tasks. LoRA, rank 64. Up to 50 turns per episode; maximum sequence length 127k tokens. STAPO~\citep{liu2026stapo} enabled; DPPO~\citep{qi2026dppo} mask with threshold 0.2; maximum staleness 5. Asynchronous training in the style of prime-rl~\citep{primeintellect2025primerl}.

\begin{figure}[H]
  \centering
  \begin{tikzpicture}[x=0.018cm,y=-0.018cm,
      nd/.style={circle,draw=gray!70,fill=white,line width=0.8pt,minimum size=8pt,inner sep=0pt},
      bp/.style={circle,draw=red!75!black,fill=red!15,line width=0.8pt,minimum size=8pt,inner sep=0pt},
      win/.style={circle,draw=red!75!black,fill=red!75!black,line width=0.8pt,minimum size=8pt,inner sep=0pt},
      lbl/.style={font=\scriptsize,text=black!70}]
    \node[bp,minimum size=10pt] (r) at (280,30) {};
    \node[nd] (a) at (150,95) {};   \node[bp] (b) at (410,95) {};
    \node[nd] (a1) at (80,160) {};  \node[nd] (a2) at (210,160) {};
    \node[nd] (b1) at (350,160) {}; \node[bp] (b2) at (470,160) {};
    \node[nd] (a21) at (170,222) {}; \node[nd] (a22) at (250,222) {};
    \node[nd] (b21) at (430,222) {}; \node[win] (b22) at (510,222) {};
    \foreach \p/\c in {r/a,a/a1,a/a2,b/b1,a2/a21,a2/a22,b2/b21}
      \draw[gray!60,line width=0.7pt] (\p) -- (\c);
    \foreach \p/\c in {r/b,b/b2,b2/b22}
      \draw[red!75!black,line width=1.3pt] (\p) -- (\c);
    \foreach \n in {a1,a21,b1}
      \draw[black!60,line width=0.8pt] ($(\n)+(-4,-4)$) -- ($(\n)+(4,4)$) ($(\n)+(4,-4)$) -- ($(\n)+(-4,4)$);
    \node[lbl,anchor=west] at (298,30) {reset scene};
    \node[lbl,anchor=west] at (428,95) {after move};
    \node[lbl,anchor=west] at (488,160) {after grasp};
    \node[font=\scriptsize\bfseries,text=red!75!black] at (510,244) {success};
    \node[lbl] at (64,186) {pruned};
    \node[lbl,anchor=west] at (20,20) {best-first path in red};
  \end{tikzpicture}
  \caption{\textbf{Prefix-tree search} (\S\ref{sec:init}). A node is created after each physical action (a move, grasp or release) settles. The search keeps returning to the most promising partial episode and branching from there, instead of starting every attempt from scratch; branches that keep failing are pruned. We allow up to about 1{,}000 leaves per task.}
  \label{fig:tree}
\end{figure}
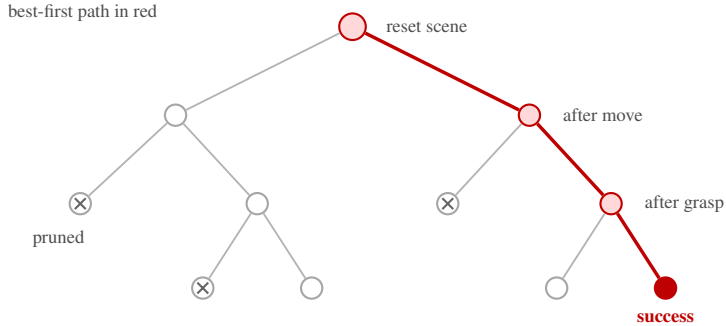

\section{Per-task real-world results}
\label{app:pertask}

\begin{table}[H]
  \centering
  \resizebox{\linewidth}{!}{%
  \begin{tabular}{@{}lccccc@{}}
    \toprule
    Task prompt & \zap{} & \astra{} & \astra{} (RC) & MolmoAct2 & Base Qwen \\
    \midrule
    Put the mustard in the left bin & 3/3 & 3/3 & 2/3 & 1/3 & 0/3 \\
    Pack canned foods into the bin & 3/3 & 2/3 & 1/3 & 1/3 & 1/3 \\
    Put the whiteboard marker in the mug & 2/3 & 3/3 & 1/3 & 0/3 & 0/3 \\
    Put the cube and the banana in the bowl & 3/3 & 3/3 & 0/3 & 2/3 & 0/3 \\
    Put the Rubiks cube in the bowl & 3/3 & 3/3 & 0/3 & 3/3 & 0/3 \\
    Put the white cup in the center of the table & 3/3 & 3/3 & 1/3 & 0/3 & 2/3 \\
    Stack the left block onto the right block & 3/3 & 3/3 & 2/3 & 0/3 & 0/3 \\
    Put the rubiks cube behind the bowl & 3/3 & 3/3 & 0/3 & 0/3 & 0/3 \\
    Put the canned food in the grey bin & 3/3 & 3/3 & 1/3 & 3/3 & 2/3 \\
    Make sure there are 3 bananas in the grey bin & 3/3 & 3/3 & 3/3 & 0/3 & 0/3 \\
    \bottomrule
  \end{tabular}}
  \caption{\textbf{Per-task successes.} RC: Robocurve harness.}
  \label{tab:pertask}
\end{table}

\section{Extended related work}
\label{sec:related}

\textbf{Compositional reasoning in robotics.} Classical robotics composes behavior explicitly. Task-and-motion planners search over symbolic operators that ground into geometric motion~\citep{toussaint2015lgp,garrett2020pddlstream,garrett2021tamp}. Learned variants acquire the operators, their abstractions or their programs from data~\citep{xu2018ntp,wang2023programport,mao2023reusable,liu2024blade}, or chain learned skills with generative models~\citep{mishra2023gsc}, or compose diffusion policies across tasks and domains~\citep{wang2024poco}. Coding agents can also synthesize generalized TAMP programs~\citep{merler2026codingtamp}. Language models later took over the high-level planner: they plan zero-shot in language or as programs~\citep{huang2022zeroshot,singh2023progprompt}, sequence a library of learned skills~\citep{ahn2022saycan,huang2022innermonologue,lin2023text2motion}, hand subgoals to RL-trained or local low-level policies~\citep{dalal2024psl,dalal2025manipgen}, or supply open-world constraints to a TAMP solver~\citep{yang2025vlmtamp,kumar2026owltamp}. \zap{} keeps the split between reasoning and motion but drops the fixed skill library. The interface is a set of tools in physical units, and motion is a skill the model learns itself, so a grasp learned on pick and place can be reused as a tube rotation or a marker stroke.

\textbf{End-to-end policies and their compositional limits.} Embodied multimodal language models~\citep{driess2023palme} and vision-language-action models (VLAs) fold reasoning and motion into one network trained on robot actions, increasingly pooled across embodiments~\citep{oxe2024openx,shukor2025smolvla,brohan2023rt2,kim2024openvla,black2024pi0,physicalintelligence2025pi05,nvidia2025gr00t,geminiroboticsteam2025gemini,fang2026molmoact2}. World-action models (WAMs) additionally predict future observations alongside actions, using video as a dense representation of how the world evolves~\citep{ye2026dreamzero,chen2026lawam,lu2026wamsurvey}. Zero-WAM uses in-context human videos to guide unseen tasks~\citep{zhou2026zerowam}, while ObjDream uses object-level representations and future object-state prediction to improve generalization to new layouts and backgrounds~\citep{ma2026objdream}. Some recover a hierarchy inside the model through embodied chain-of-thought or a high-level policy~\citep{zawalski2024ecot,shi2025hirobot,huang2025thinkact,geminiroboticsteam2025gemini15}. Several studies show that these policies keep the intent of an instruction but fail to execute it on new tasks, held-out perturbations or new skill combinations~\citep{zhou2025agnostos,fang2025intact,zhou2025liberopro,yang2026craft}. Recent diagnoses also find that current VLAs struggle with instructions specifying how to carry out a task~\citep{park2026instructions}. One reason is that compositional coverage in action data is expensive to collect~\citep{gao2024compositional}. Benchmarks now measure compositional generalization directly, from compositional RL suites~\citep{mendez2022composuite} to arm-wise composition in dual-arm VLAs~\citep{zhang2026acgbench}, alongside standard manipulation benchmarks~\citep{liu2023libero} and real-to-sim evaluation~\citep{li2024simplerenv}. \zap{}'s training tasks are deliberately simple and teach it only how to move, entirely in simulation; the reasoning it uses on tasks comes from pretraining.

\textbf{Foundation models as robot agents.} Code as Policies and its successors compose perception and control APIs into programs or value maps~\citep{liang2022code,huang2023voxposer,huang2024rekep}, in the style of tool-using agents~\citep{yao2023react}. RAX compiles a coding agent's online exploration into robot programs that run without the agent~\citep{huang2026rax}; MoCaP uses passive human videos to guide code-as-policy for mobile manipulation~\citep{cai2026mocap}. Multimodal language-model agents remain weak at low-level manipulation when used off the shelf~\citep{yang2025embodiedbench}. Recent work nonetheless uses frontier agents directly as closed-loop policies: on real YAM arms~\citep{jia2026agentpolicy,robocurve2026inspect,menon2026astraarms}, including head-to-head comparisons with VLAs on bimanual tasks~\citep{zhang2026stationerybench} and in simulation~\citep{cookiegg2026robodojo}, through general agent frameworks~\citep{tsui2026faea,chu2026openrua}, through a visual interface~\citep{hu2026via}, through in-context learning~\citep{cheng2026incontext}, through primitive motion commands~\citep{guo2026robodawn}, or with perception tools and test-time simulation~\citep{li2026roboharness,zhang2026worldactionagent,liu2025simpact}. SEAR compares code-as-policy, agent-as-policy and VLA auto-research, with controls that separate retained experience from repeated attempts and larger budgets~\citep{wang2026sear}.

Other work improves the harness, with evolving code, skill libraries or playbooks~\citep{wang2026harnesspai,lu2026aspire,kim2026rhd,shen2026embodiedswe,liu2026rapid,wang2026dexagent}. URAI lets frontier agents write, call and evolve reusable robot tools~\citep{ge2026urai}, while ChunkAct chains primitives into action chunks to reduce model calls~\citep{xu2026chunkact}. Some harnesses surround a VLA: \emph{Cracking Mobile Manipulation} adds planning, navigation and local-motion interventions around a fixed policy~\citep{wang2026mobilemanipulation}, and CarryOn adds planning and execution loops with reusable recovery skills~\citep{wang2026carryon}. AdaRSI first improves the harness, then fine-tunes the executor on sub-steps that remain unreliable~\citep{wang2026adarsi}.

Concurrent work adapts frontier-agent skills from simulation to hardware without weight updates~\citep{wang2026rpg,he2026skill2real,li2026robobridge}. SimEX develops a robot toolbox in simulation, then adapts it and the simulator through a few physical trials~\citep{hu2026simex}. \zap{} trains the motion skill into an open 27B model that runs 3$\times$ faster and 17$\times$ cheaper than the frontier model it matches.

\textbf{VLMs that directly predict actions.} EASG projects a tool-center-point action frame into the wrist-camera image and lets a pretrained VLM generate continuous translations and rotations as text~\citep{yi2026easg}. This grounds the action space for zero-shot manipulation, and the same interface supports fine-tuning on robot trajectories. \zap{} also uses the model's native text generation to command motion, learning full absolute end-effector poses through SFT and RL entirely in simulation.

\textbf{Small trained robot agents.} The closest work trains compact agents. Guava distills frontier-model trajectories collected in simulation into a 4B agent using semantic grasp and align tools~\citep{liu2026guava}. RACaP evolves reusable policy APIs, then distills frontier-model ReAct decisions into an 8B model for efficient deployment~\citep{li2026racap}. Show-Harness fine-tunes a 2B model on discrete action units, partly on RoboLab~\citep{chen2026showharness}. Robo-Harness K1 distills a 9B student, evaluated only in simulation~\citep{li2026roboharness}. CaP-RL trains a 7B code-writing model with GRPO in simulation and transfers it to cube lifting and stacking~\citep{fu2026capx}.

Other work fine-tunes VLM agents for manipulation question answering~\citep{bhat2026probe} or calls a small agent-ready action model through tools~\citep{jiang2026arwam}. \zap{} trains mainly with RL on its own multi-turn simulation rollouts to command full robot poses. Beyond Guava's tested recombinations of familiar skills, \zap{} shows new real-world strategies, including a mid-air handoff and marker drawing, and transfers the same checkpoint from bimanual YAM to simulated Franka without retraining.

\textbf{RL for language-model agents and embodied reasoning.} Online RL can ground language models as policies in text and simulated environments~\citep{carta2023glam,szot2024llarp,zhai2024rlvlm}. Multi-turn agent RL has since scaled with GRPO-style objectives~\citep{shao2024deepseekmath,pikus2025hardexamples,zhou2024archer,wang2025ragen,wang2025vagen}. In robotics, RL has sharpened embodied and physical reasoning in VLMs, for example pointing and keypoint prediction~\citep{yuan2026embodiedr1,kim2025robotr1,nvidia2025cosmosreason1}, has fine-tuned VLAs in simulation~\citep{liu2025rl4vla,li2025simplevlarl,lu2025vlarl}, and has trained agentic recovery decisions over VLA execution~\citep{zhang2026agenticrl}, including at scale for sim-to-real transfer~\citep{choi2026scalingvlarl}. \zap{} applies multi-turn GRPO to a VLM tool agent whose actions are real robot motions.

\textbf{Sim-to-real.} Domain and dynamics randomization made simulation-only training viable for low-level control~\citep{tobin2017domain,peng2018dynamics,openai2019rubik}. Large procedurally generated simulations now transfer manipulation policies zero-shot~\citep{deshpande2026molmobot}, and object-centric abstractions narrow the gap further~\citep{kim2026objectresidual}. We use Isaac Sim and RoboLab~\citep{nvidia_isaacsim,nvidia2025isaaclab,yang2026robolab}. Our tool interface acts as the abstraction: the observations and actions the model sees are the same in simulation and on hardware~\citep{liu2026simrealgap}. As a result, only dynamics and perception, not the interface, differ between the two.

\end{document}